\pdfoutput=1

\documentclass[11pt]{article}

\usepackage[]{acl}

\usepackage{inconsolata}

\usepackage{times}
\usepackage{latexsym}
\usepackage{beamerarticle}
\usepackage{latexsym}
\usepackage{microtype}
\usepackage{booktabs}
\usepackage{url}
\usepackage{eurosym}
\usepackage{todonotes}
\usepackage{multirow}
\usepackage{subcaption}
\usepackage{amssymb}
\usepackage{amsmath}
\usepackage{mathtools}
\usepackage{array}
\usepackage{pgfplotstable}
\usepackage{pgfplots}
\usepackage{soul}
\usepackage{xcolor}
\usepackage{txfonts}
\usepackage{pgf}
\usepackage{tikz}
\usepackage{adjustbox}
\usepackage{appendix}
\usepackage{enumitem}
\usepackage{longtable}
\usepackage{braket}
\usepackage{makecell}
\usepackage{xcolor}
\usepackage{graphicx}
\usetikzlibrary{shapes.arrows,shadings}

\definecolor{lightsalmon}{HTML}{FAE1DD}
\definecolor{darksalmon}{HTML}{fec5bb}
\definecolor{lightbeige}{HTML}{FFE6CC}
\definecolor{lightgreen}{HTML}{d8e2dc}
\definecolor{darkorange}{rgb}{0.99,0.67,0.3}
\definecolor{mediumgreen}{rgb}{0.6,0.88,0.74}
\definecolor{lightteal}{rgb}{0.6,0.77,0.74}
\definecolor{lightmagenta}{rgb}{0.84,0,0.45}

\usepackage[T1]{fontenc}

\usepackage[utf8]{inputenc}

\usepackage{microtype}

\title{To Adapt or to Annotate: \\
Challenges and Interventions for Domain Adaptation in \\
Open-Domain Question Answering}

\author{Dheeru Dua$^1$\thanks{*This work was done when the first author was
an intern at Google Research.} \quad Emma Strubell$^{2,3}$ \quad Sameer Singh$^1$ \quad Pat Verga$^2$ \\
    $^1$University of California Irvine \quad $^2$ Google Research \quad $^3$ Carnegie Melon University\\
	}

\begin{document}
\maketitle
\begin{abstract}
Recent advances in open-domain question answering (ODQA) have demonstrated impressive accuracy on standard Wikipedia style benchmarks. However, it is less clear how robust these models are and how well they perform when applied to real-world applications in drastically different domains. While there has been some work investigating how well ODQA models perform when tested for out-of-domain (OOD) generalization, these studies have been conducted only under conservative shifts in data distribution and typically focus on a single component (ie. retrieval) rather than an end-to-end system. In response, we propose a more realistic and challenging domain shift evaluation setting and, through extensive experiments, study end-to-end model performance. We find that not only do models fail to generalize, but high retrieval scores often still yield poor answer prediction accuracy. We then categorize different types of shifts and propose techniques that, when presented with a new dataset, predict if intervention methods are likely to be successful. Finally, using insights from this analysis, we propose and evaluate several intervention methods which improve end-to-end answer F1 score by up to $\sim$24 points.

\end{abstract}


\section{Introduction}

General-purpose open-domain question answering~\cite{chen2017reading,lee2019latent} is an important task that necessitates reading and understanding a large number of documents and succinctly answering a given question. It is especially crucial in fields such as Biomedicine, Legal, News, etc., where a huge number of documents are added everyday and domain expertise is necessary to understand these documents.

\begin{figure}[htpb]
    \centering
    \includegraphics[width=0.45\textwidth]{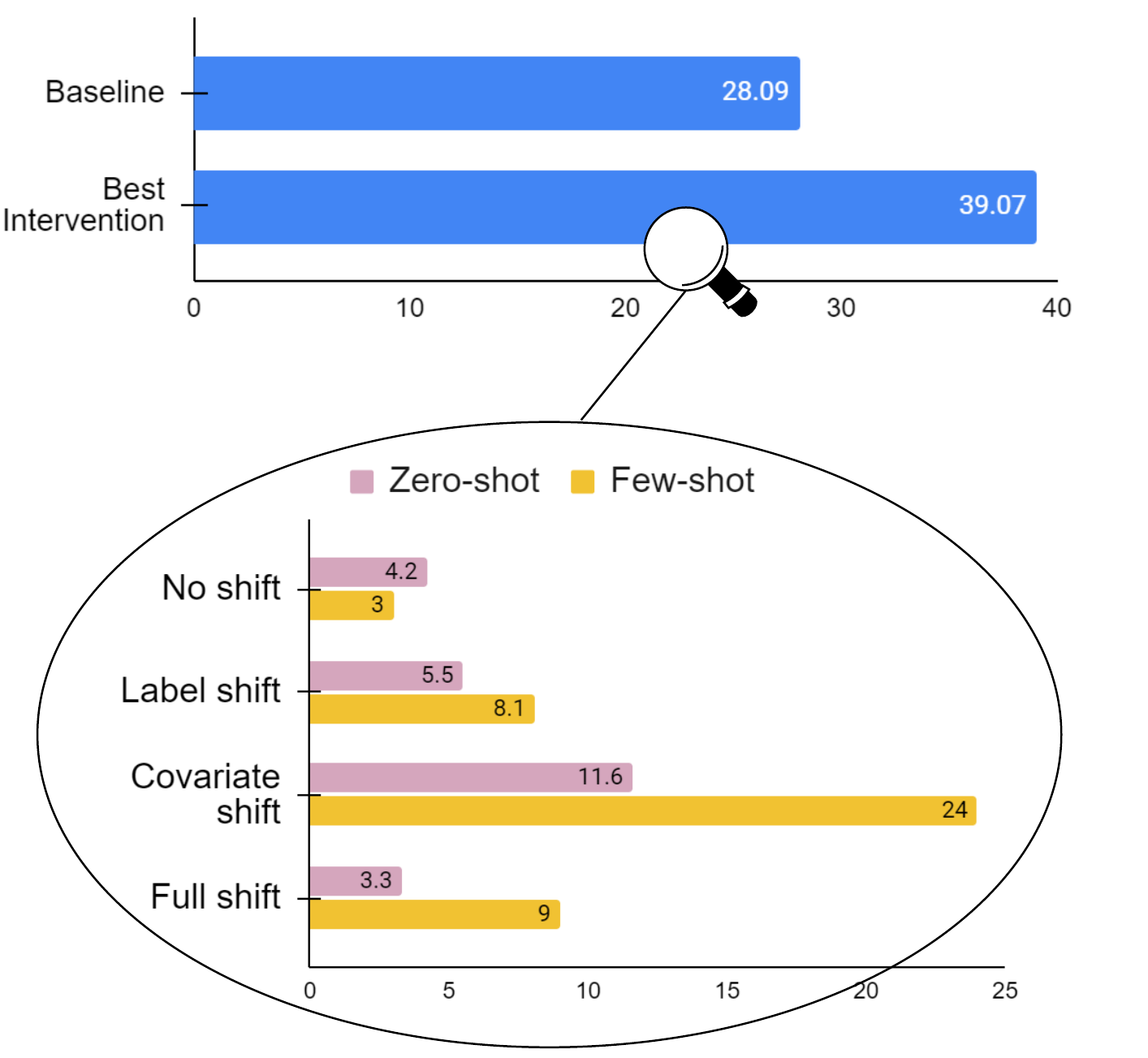}
    \caption{\emph{Top:} Average Reader performance for Baseline and best interventional or augmentation setup on top. \emph{Bottom:} The difference between baseline and performance (end-to-end F1) after introducing interventions, averaged over datasets exhibiting specific shift types.}
    \label{fig:delta}
\end{figure}

Recently, there have been great advancements and successes in open-domain question answering models (ODQA). The state of the art ODQA systems perform a two-stage pipeline process~\cite{izacard2022few}: 1) given a question, a context \emph{retriever}~\cite{karpukhin2020dense,izacard2021unsupervised,raffel2020exploring} selects relevant passages and 2) a question answering model, also known as \emph{reader}~\cite{izacard2020leveraging} answers the given question based on the retrieved passages. This decoupling allows for independent advancements in domain generalization and adaptation of general-purpose context retrievers~\cite{thakur2021beir} and question answering~\cite{fisch2019mrqa} models. 

A general purpose ODQA model should be resilient to changes in document, question and answer distributions. However, existing works seldom study the effectiveness of a model trained on a particular source domain and applied to a new target domain. In this work, we ask the following questions:


\begin{enumerate}
    \item How well do current state-of-the-art ODQA methods perform when tested on varying degrees of data shift and under what conditions they fail?
     \item Given a small set of labeled examples in the new target domain, can we predict whether existing intervention schemes would be useful in adapting from a given source model or would it better to collect annotations in the target domain?
    \item What interventions or adaptation strategies can we perform to improve ODQA performance in OOD testing? 
   
\end{enumerate}

Following the above research questions we make four primary contribution in this work.

First, in Section~\ref{sec:setup} we aggregate a set of seven ODQA datasets, spanning five different domains for evaluating domain generalization of an ODQA model trained on general purpose. In Section~\ref{sec:zero_shot_gen}, we use this test-bed to show that most SotA ODQA models fail to generalize, and go on to analyze the failure modes for OOD generalization. For example, we observe that the retriever model's performance is quite sensitive to the type of entities and length of passages seen at training, and additionally, in $\sim65\%$ of cases where the answer string appears in the retrieval list (one of the most commonly used metrics for retrieval accuracy), the context does not justify the answer.

Second, in Section~\ref{sec:shift}, we propose a generalizability test, that determines the type of dataset shift in the target datasets with only a few labeled examples in target domain.  This gives us an idea of how likely it is for a model trained in the source domain to adapt to an unseen target domain. In Figure~\ref{fig:delta}, we observe that target datasets which are close to source domain and exhibit `No shift', do not show much improvement with zero or few shot data augmentation. While the target datasets, that are very different from source data and exhibit `Full shift' need examples generated in a few shot way that capture the underlying target domain to adapt to the target dataset. We consider few-shot examples as proxy for target data distribution. Zero-shot data augmentation techniques yield best adaptation under `Label shit' and `Covariate shift'. 

Third, in Section~\ref{sec:zero_shot_adapt}, we analyze the performance impact of various intervention schemes, such as heuristic data augmentations and language model generated pseudo-data, without relying on any labeled target data. We observe that zero-shot data interventions yield a relatively high improvement in performance (up to 15\% in F1) for some shift types, while others do not see these gains. 


Finally, in Section~\ref{sec:few_shot}, we propose a simple and effective technique for few-shot language model data generation requiring only a handful of examples from the target domain. While many existing works have leveraged question generation models for creating additional training data, these models are typically trained on the source domain data and suffer the same generalization shortcomings. Instead, inspired by the strong performance of large language models (LLM) for summarization, we generate sentences by prompting the LLM with a handful of examples from the target domain. We convert the generated sentences into cloze style QA pairs and show that this technique is especially effective when zero shot adaptation methods fail to capture the target domain distribution, yielding improvements of up to 24\% in F1.

\section{Background and Setup}
\label{sec:setup}
An open-domain (ODQA) model learns interactions among three random variables: question ($\mathbb{Q}$), answer ($\mathbb{A}$) and context ($\mathbb{C}$). For a given $q \in \mathbb{Q}$, first the retriever $\mathcal{R}$ returns a set of passages, $c_q = \mathcal{R}(q, \mathbb{C})$. These passages are then sent to an answering model $\mathcal{M}$ (also known as reader) to obtain the final answer, $\hat{a} \leftarrow \mathcal{M}(a| q, c_q)$. 

In our experiments, we follow prior work and compute retriever performance as Accuracy at K(Acc@k), which computes if the oracle answer is found in the top-$k$ retrieved passages\footnote{The only exception is he COLIEE dataset which primarily contains boolean (yes/no) answers so we instead use oracle passages to compute Acc@100}. We set  $k$=100 in all of our experiments. To measure the reader performance, we compute token-level $F_1$ between the oracle answer and prediction from the answering model\footnote{We do not consider the other common reader metric of exact-match to reduce the occurrences of minor dataset annotation guidelines leading to a 0 score for a reasonable answer.}.

\subsection{Datasets}
In this work, we test the generalizability of a model trained on a \emph{source domain} to seven datasets in five vastly different \emph{target domains}.

\noindent{\textbf{Source Domain:}}
For all of our experiments, our source domain is English Wikipedia along with the supervised data from NaturalQuestions (NQ)~\cite{kwiatkowski2019natural} and BoolQ~\cite{clark2019boolq}. We treat this domain as our source as it used for the vast majority of current work in ODQA (and many other areas of language research). 

In addition to the supervised training data from NQ and BoolQ, we add additional cloze style questions derived from the QA pairs in NQ. For each qa pair, we retrieve a sentence from Wikipeida with the highest BM25 similarity score. We then convert the retrieved sentence into a cloze-style question by replacing the answer string in the sentence with sentinel markers~\cite{raffel2020exploring}\footnote{We use cloze augmentation for training reader models because some target datasets contain cloze-style questions, keeping the question distribution consistent across different experimental setups. We do not perform this augmentation for retrievers because we observed a performance drop. 
}.

\noindent{\textbf{Target Domains:}}
We consider five vastly different domains (Stackoverflow, Reddit, Pubmed, Japanese Statute Law codes, CNN/Dailymail and Wikipedia) as our target corpora and re-purpose seven open-domain QA and/or reading comprehension datasets for our evaluations (Figure~\ref{tab:ds}). The datasets are Quasar-S~\cite{dhingra2017quasar}, Quasar-T~\cite{dhingra2017quasar}, SearchQA~\cite{dunn2017searchqa} and BioASQ~\cite{balikas2015bioasq} which were introduced as ODQA datasets over Stackoverflow, Reddit, Wikipedia and Pubmed corpus respectively.

Additionally, we re-purpose NewsQA~\cite{trischler2016newsqa} and CliCR~\cite{vsuster2018clicr} as ODQA datasets. These datasets were originally introduced as reading comprehension evaluations and constructed by retrieving a set of passages for the given question from Pubmed and CNN/Dailymail corpus. We also re-purpose COLIEE~\cite{rabelo2022overview}, originally an entailment based QA dataset, by transforming the examples into boolean questions and retrieving passages from a Japanese Statute Law corpus. End-to-end performance of ODQA models trained on target QA pairs with BM25 retrievals from the target corpus (UB-Ret, Figure~\ref{fig:reader_zero_shot}), indicates that these datasets can be reasonably re-purposed for our ODQA setup. Figure~\ref{fig:evaluate_set} shows some examples from target datasets.

\begin{figure*}
\footnotesize
\begin{center}
\begin{tabular}{p{1.5cm}p{1cm}p{8cm}p{5cm}}
    \toprule
      {\bf Dataset, Corpus} & {\bf \#ques, \#docs} & {\bf Passage} & {\bf Question-Answer}\\
      \midrule
      BioASQ, Pubmed & 5k, 30M & Parkinson's disease (PD) is one of the most common degenerative disorders of the central nervous system that produces motor and non-motor symptoms. The majority of cases are idiopathic and characterized by the presence of Lewy bodies. & Q: Which disease of the central nervous system is characterized by the presence of Lewy bodies? A:  Parkinson's disease \\
      \hline
      CliCR, Pubmed & 90k, 30M & Detailed history and examination ruled out the above causes except the exposure to high altitude as a cause for koilonychia in our patient. Exposure to high altitude is a known aetiology for koilonychias, also described by some authors as ``Ladakhi koilonychia''. & Q: \_\_ is a known cause of koilonychia, described by some as Ladakhi koilonychia. A: High altitude exposure \\
      \hline
      Quasar-S,  Stackoverflow & 30k, 1.5M & I have a mixed integer quadratic program MIQP which I would like to solve using SCIP. The program is in the form such that on fixing the integer variables the problem turns out to be a linear program. & Q: scip -- an software package for solving mixed integer \_\_  problems A: linear-programming\\
      \hline
      Quasar-T, Reddit & 30k, 2M & Because of widespread immunization , tetanus is now rare. Another name for tetanus is lockjaw. & Q: Lockjaw is another name for which disease A: tetanus \\
      \hline
      NewsQA, Dailymail & 70k, 0.5M & Former boxing champion Vernon Forrest, 38, was shot and killed in southwest Atlanta, Georgia, on July 25.
 & Q: Where was Forrest killed ? A: in southwest Atlanta , Georgia
\\
      \hline
      SearchQA, WIkipedia & 70k, 20M & The Dangerous Summer and The Garden of Eden. Written in 1959 while Hemingway was in Spain on commission for Life...
 & Q: While he was in Spain in 1959, he wrote “The Dangerous Summer”, a story about rival bullfighters A: Hemingway \\
      \hline
      COLIEE, Japanese Legal Codes & 886, 1k & 
 A manifestation of intention based on fraud or duress is voidable. If a third party commits a fraud inducing a first party to make a manifestation of intention to a second party, that manifestation of intention is voidable only if the second party knew or could have known that fact. The rescission of a manifestation of intention induced by fraud under the provisions of the preceding two paragraphs may not be duly asserted against a third party in good faith acting without negligence. & Q: Is it true: A person who made a manifestation of intention which was induced by duress emanated from a third party may rescind such manifestation of intention on the basis of duress, only if the other party knew or was negligent of such fact. A: No \\
      \bottomrule
\end{tabular}
\end{center}
\caption{Examples from datasets with context and question-answer pairs from different domains. }
\label{fig:evaluate_set}
\end{figure*}

\subsection{Models}
\noindent{\textbf{Retrievers:}} We compare four diverse retrievers: 1) BM25~\cite{robertson1994simple} (sparse and unsupervised), 2) Contriever (semi-supervised with MS-MARCO) ~\cite{izacard2021unsupervised} 3) Dense Passage Retriever (DPR) ~\cite{karpukhin2020dense} and the state-of-the-art model 4) Spider~\cite{ram2021learning} (supervised with NaturalQuestions). 
 
\noindent{\textbf{Reader:}} We use the state-of-the-art T5-large based fusion-in-decoder (FiD) model \citep{izacard2020leveraging} which encodes top 100 documents in parallel. The representation are concatenated and then decoded to generate the final answer.

\section{Categorizing Data Shift Types}
There are many aspects that determine in what ways and to what extent one data distribution differs from another. Having a better understanding of this spectrum of possibilities would enable us to predict whether a new dataset would be compatible with an existing model and, if not, what types of interventions would be required in order to enable the model to adapt to the new domain.

In this section, we define a taxonomy of shift types for ODQA based on the distributions of the sub-components of the problem (answer, question, and context distributions), and develop of technique for categorizing target domains amongst those shift types. While we find in later sections that the type of shift often influences the effectiveness of an interventional scheme (See sections \ref{sec:zero_shot_interventions} and \ref{sec:few_shot}), we also find that actually determining the type of shift is quite challenging.

\label{sec:shift}
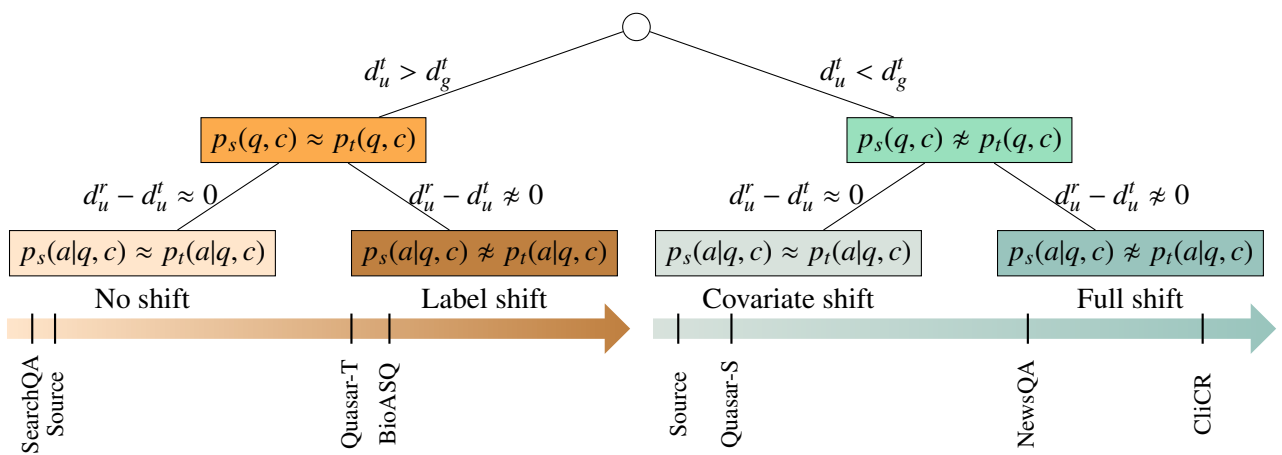
\begin{figure*}
\centering
    \begin{tikzpicture}[level 1/.style={sibling distance=85mm},level 2/.style={sibling distance=45mm,level 3/.style ={level distance=0.6cm}}]
    \node[circle, draw] {}
      child {node [rectangle, draw, fill=darkorange] {$p_s(q,c) \approx p_t(q,c)$}
        child {
          node[rectangle, draw, fill=lightbeige] {$p_s(a|q,c) \approx p_t(a|q,c)$}
          child { [edge from parent/.style={draw=none}] node[draw=none] {No shift}}
          edge from parent  node[left] {$d_u^r - d_u^t \approx 0$} 
          }
        child {
          node[rectangle, draw, fill=brown] {$p_s(a|q,c) \not\approx p_t(a|q,c)$}
          child { [edge from parent/.style={draw=none}] node[draw=none] {Label shift}}
          edge from parent  node[right] {$d_u^r - d_u^t \not\approx 0$} 
         }
        edge from parent node[left, xshift=-0.5cm] {$d_u^t > d_g^t$ };
        \node[single arrow, left color=lightbeige, right color=brown, minimum width=5mm, single arrow head extend=2mm, minimum height=82mm, yshift=-2.5cm] {};
        \node[draw=none,text=black,yshift=-3.5cm,xshift=-3.7cm,rotate=90]{\footnotesize{SearchQA}};
        \node[draw=none,text=black,yshift=-3.5cm,xshift=-3.4cm,rotate=90] {\footnotesize{Source}};
        \node[draw=none,text=black,yshift=-3.5cm,xshift=0.5cm,rotate=90] {\footnotesize{Quasar-T}};
        \node[draw=none,text=black,yshift=-3.5cm,xshift=1.0cm,rotate=90] {\footnotesize{BioASQ}};
        \draw [thick] (-3.7,-2.3) -- (-3.7,-2.7);
        \draw [thick] (-3.4,-2.3) -- (-3.4,-2.7);
        \draw [thick] (0.5,-2.3) -- (0.5,-2.7);
        \draw [thick] (1.0,-2.3) -- (1.0,-2.7);
      }
      child {node [rectangle, draw, fill=mediumgreen] {$p_s(q,c) \not\approx p_t(q,c)$}
        child {
           node[rectangle, draw, fill=lightgreen] {$p_s(a|q,c) \approx p_t(a|q,c)$}
           child { [edge from parent/.style={draw=none}] node[draw=none] {Covariate shift}}
           edge from parent  node[left] {$d_u^r - d_u^t \approx 0$ } 
         }
        child {
          node[rectangle, draw, fill=lightteal] {$p_s(a|q,c) \not\approx p_t(a|q,c)$}
          child { [edge from parent/.style={draw=none}] node[draw=none] {Full shift}}
          edge from parent  node[right] {$d_u^r - d_u^t \not\approx 0$} 
         }
      edge from parent node[right, xshift=0.5cm] {$d_u^t < d_g^t$ };
      \node[single arrow, left color=lightgreen, right color=lightteal,
      minimum width=5mm, single arrow head extend=2mm,
      minimum height=82mm, yshift=-2.5cm] {};
      \node[draw=none,text=black,yshift=-3.5cm,xshift=-3.7cm,rotate=90]{\footnotesize{Source}};
      \node[draw=none,text=black,yshift=-3.5cm,xshift=-3.0cm,rotate=90] {\footnotesize{Quasar-S}};%
      \node[draw=none,text=black,yshift=-3.5cm,xshift=0.9cm,rotate=90] {\footnotesize{NewsQA}};
    \node[draw=none,text=black,yshift=-3.5cm,xshift=3.2cm,rotate=90] {\footnotesize{CliCR}};
    \draw [thick] (-3.7,-2.3) -- (-3.7,-2.7);
    \draw [thick] (-3.0,-2.3) -- (-3.0,-2.7);
    \draw [thick] (0.9,-2.3) -- (0.9,-2.7);
    \draw [thick] (3.2,-2.3) -- (3.2,-2.7);
      };
    \end{tikzpicture}
    \caption{Generalizability test: At first level, the farther the target distribution from uniform as compared to gold, the closer it is to the source. At second level, the gradual increase from left to right in the leaf nodes depicts increase in difference between distance of reference (source) from uniform and distance of target from uniform. The lower the difference (i.e, the left branch at final depth), the closer is the target to source.}
    \label{fig:shift}
\end{figure*}

\subsection{Types of dataset shift}
Each domain contains both an input distribution (questions, contexts) and output distribution (answers). The compatibility - or lack-there-of - over these two sub-distributions lead to four possible settings.
\paragraph{No shift} both the input and output distributions between the source and target domain match.
\paragraph{Label shift~\cite{storkey2009training}} occurs when the input distributions of the source and target domains match, i.e., $p_s(x_s) = p_t(x_t)$ while the output label distribution given the input between source and target domain does not match, $p_s(y_s|x_s) \neq p_t(y_t|x_t)$. 

\paragraph{Covariate shift~\cite{zadrozny2004learning}} occurs when the source and target input distributions do not match i.e., 
$p_s(x_s) \neq p_t(x_t)$ while the output label distributions matches $p_s(y_s|x_s) = p_t(y_t|x_t)$.

\paragraph{Full shift} occurs when both the source and target input and output distributions do not match.

\subsection{Calculating Shift in ODQA}
Most existing works consider classification setups where it is easy to compute the input and output distributions. However, in our setting, we lack a consistent method for computing these distributions which often require large amounts of labeled target data to train a target model for comparison. As an alternative, we determine the type of dataset shift by estimating whether the source model contains useful information about the input and output distributions of the target dataset when compared with an uninformative uniform prior.

We characterize shift in ODQA as a two-step process. We first compute the input distribution, i.e, the joint question and context distribution using unnormalized (energy) scores from a dense retriever~\cite{karpukhin2020dense} to quantify the compatibility between a given question and a context via $\mathcal{R}(q,c)$. Then, obtain the likelihood of the gold context for a given question by normalizing the energy scores from the retriever over a set of contexts. This computation over the entire corpus can be very expensive and results in a low entropy distribution. To address this, we sample a set of contexts, $\mathcal{C}$, from the entire corpus $\mathbb{C}$. 
\begin{equation}
    p(q, c_g) = \frac{\mathcal{R}(q,c_g)}{\sum_{c_k \in \mathcal{C}} \mathcal{R}(q,c_k)} 
    \label{eq:ret}
\end{equation}

In the second step, we test if the output distributions match by computing the likelihood of generating an oracle answer given a question and the relevant contexts. We use global normalization~\cite{goyal2019empirical} for computing the probability distribution over a set of answer spans. Ideally, the normalization should be computed over all possible answer spans in the corpus which is intractable. We instead sample a set of answer spans to approximate the normalizer.

\begin{equation}
    p(a_g|q, c_q) = \frac{\prod_{t}\mathcal{M}(a_g^t|a_g^{<t}, q,c_q)}{\sum\limits_{a_k \in \mathcal{A}} \prod_{t} \mathcal{M}(a_k^t|a_k^{<t}, q,c_q)}
    \label{eq:read}
\end{equation}

\subsection{Predicting type of dataset shift}
Adapting or fine-tuning a pre-trained source model to match the target domain, can be formulated in a Bayesian framework. The source model acts as a prior which when exposed to interventional data, that estimates the likelihood of target domain, results in a (fine-tuned) posterior distribution.
\begin{align*}
    \underbrace{p(\theta_t|x_t)}_{posterior} &= \underbrace{p(\theta_s; x_s)}_{prior} \underbrace{p(x_t|\theta_s; x_s)}_{likelihood} \\
    &= p_s(x_s) p_t(x_t|x_s)
\end{align*}

To analyze the type of dataset shift, we devise a 
\emph{generalizability test}, where we compare the prior distribution to an uninformative prior like the uniform distribution. In particular, if the source model is closer to the uniform distribution when compared with the oracle  distribution it does not have the reasoning  ability (informative signal) to understand the target domain. We assume we have access to a few labeled examples in target domain for evaluation.  

\paragraph{Input/Retriever Distribution:} In the first stage, we compute the input distribution using retriever scores by following Eq.~\ref{eq:ret}. Then, for a given question, we compute the distance of the input distribution of the target domain  from the uniform distribution, $d_u^t$ and average the distances over the set of examples in the evaluation set. Similarly, we also compute the distance from the gold distribution as $d_g^t$. If $d_u^t > d_u^g$, we conclude that the distance of a target distribution is far from the uniform distribution and closer to the gold distribution, indicating that the source distribution is likely compatible with the target domain (Figure~\ref{fig:shift}). Since we do not assume access to labeled target domain data for training, this compatibility measure is used as a proxy to infer that $p_s(q,c) \approx p_t(q,c_t)$. We use this notation as a way to interpret that source and target distributions are compatible and not necessarily equal.

\begin{table}[!htb]
\centering
    \small
    \begin{tabular}{lccl}
    \toprule
      Dataset & {\bf Retriever}  & \bf{Reader} & {Shift} \\ 
     \midrule
      BioASQ & \textcolor{brown}{0.3027} &  0.1765 & Label \\ 
      CliCR & \textcolor{teal}{-0.8839} &  0.2352 & Full\\
      Quasar-S & \textcolor{teal}{-0.6697}  &  0.0767 & Covar. \\
      Quasar-T & \textcolor{brown}{0.2016}  & 0.1694 & Label\\ 
      NewsQA  & \textcolor{teal}{-0.1967}  & 0.1800 & Full\\ 
      SearchQA & \textcolor{brown}{0.6165}  & -0.0063 & No\\ 
    \bottomrule
    \end{tabular}
    \caption{Wasserstein distance: Computed over 100 examples labeled examples from target domain. The reference of source domain model has $d_u^r$=0.2925}
    \label{tab:ret_wd}
\end{table}

\paragraph{Output/Reader Distribution:} In the second stage, we follow a similar procedure to characterize for the output distribution. To analyze the compatibility between the output answer distribution and a uniform distribution, we need to compute a probability distribution over a set of answers similar to stage 1. However,  the conditional answer generation model is not trained with a contrastive loss like the retriever leading to the answer likelihood distribution having a higher entropy. Also, the support set of answers used for normalization contains only grammatically correct answer spans making the likelihood scores attenuated. To deal with these issues, we use a reference answer conditional distribution to de-bias the likelihood scores with a threshold. We treat the source distribution as our reference and compute the distance from the uniform and gold distributions with respect to the source distribution on 100 examples from the validation set of the source domain. To infer if $p_s(a|q,c) \approx p_t(a|q,c)$, we determine the difference between the distance of reference distribution from uniform and distance of target distribution from uniform. If this difference is close to 0, we conjecture that the $p_s(a|q,c)$ and $p_t(a|q,c)$ are compatible.

In Figure~\ref{fig:shift}, we can see that the dataset SearchQA falls under the ``No shift" category, hence, we conjecture that it will observe minimal improvements under most data intervention schemes, as the source is already able to capture the target distribution well (Section~\ref{sec:zero_shot_adapt},~\ref{sec:few_shot}). We also conjecture that datasets falling under the category of ``Label shift'' and ``Covariate shift'' are more amenable to zero-shot data interventions, however, ``Full shift'' would benefit most from few-shot examples or collecting annotations in the target domain.
We consider few shot augmentations as a proxy for annotating examples in the target domain because the augmentations are generated with supervision from target data. 

\section{How Well do Models Generalize? \label{sec:zero_shot_gen}}
In this section, we want to first get a sense of how well existing SotA ODQA models perform when tested OOD.
We test the OOD performance of source-trained models on target domain validation sets and, when they fail, analyze what caused those errors.

\subsection{End-to-End Zero-shot Generalization} 
in Figure~\ref{fig:reader_zero_shot}, we test the end-to-end domain adaption performance of three model variants:

\noindent\textbf{Source:} a fully source domain trained model with BM25 retrieved documents, demonstrating zero-shot generalization performance.

\noindent\textbf{Upperbound-Reader} a target domain trained reader model with contexts retrieved by BM25 -- the overall strongest retriever.

\noindent\textbf{Upperbound-Retriever} a target domain trained reader model with gold contexts to approximate upper-bound performance.

Overall, when testing models on the new target domains we observe large performance drops. This is especially true when the target corpus differs from Wikipedia, such as in Quasar-S (stackoverflow) and CliCR (pubmed), even though the model requires similar reading capabilities to those needed in the source domain.   

Interestingly, even though the BM25 retriever accuracy is relatively high on the target datasets, (for example, $\sim$83\% Acc@100 on Quasar-S), that accuracy does not translate to strong reader performance and therefore, overall QA accuracy ($\sim$11\% F1 on Quasar-S, Figure~\ref{fig:reader_zero_shot}). 

To understand the performance gap, we manually sample 50 prediction from each target dataset where retrieved passages contain the oracle answer but the reader produced an incorrect prediction. We observe that in \textbf{around 65\% cases, the Acc@100 metric yields a false positive}, where the passage contains an exact string match of the correct answer, but the context does not actually answer the given question.
For example, the question ``What is the name of the office used by the president in the white house?" and answer ``oval", the retrieved passage ``A tunnel was dug into the White House connecting the Oval Office to a location in the East Wing...." is credited to able to answer the question. This shows that end-to-end performance is crucial in understanding improvements in retrievers which is often ignored.

\pgfplotsset{compat=1.11,
    /pgfplots/xbar legend/.style={
    /pgfplots/legend image code/.code={%
       \draw[##1,/tikz/.cd, rotate=90,yshift=-0.25em]
        (0cm,0cm) rectangle (3pt,1.5em);},
   },
}
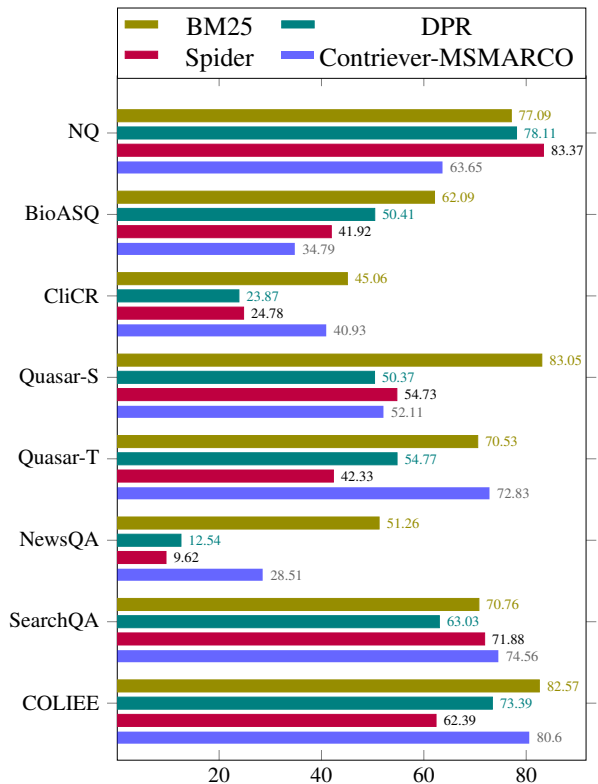
\begin{figure}[ht]
    \small
    \pgfplotstableread[col sep=space]{plots/retriever-zero-shot.dat}{\retrievertable}
    \centering 
    \begin{tikzpicture}[scale=0.9]
    \begin{axis}[xbar,
      y=1.2cm,
      xmin=0.1,
      legend columns=2,
      xlabel=,
      ytick=data,
      nodes near coords,
      every node near coord/.append style={font=\tiny},
      yticklabels from table={\retrievertable}{dataset},
      legend style={at={(0.5,1.1)},anchor=north,font=\normalsize,minimum width=1.8cm}
      ]
    
    \pgfplotsinvokeforeach{BM25}{
        \addplot [bar width=0.07in,bar shift=0.1in,fill=olive,draw=olive, color=olive]
            table [y expr=\coordindex, x=#1] {\retrievertable};
        \addlegendentry{#1};
    }
    \pgfplotsinvokeforeach{DPR}{
        \addplot [bar width=0.07in,bar shift=0.0001in,fill=teal,draw=teal,color=teal]
            table [y expr=\coordindex, x=#1] {\retrievertable};
        \addlegendentry{#1};
    }
    \pgfplotsinvokeforeach{Spider}{
        \addplot [bar width=0.07in,bar shift=-0.1in,fill=purple,draw=purple]
            table [y expr=\coordindex, x=#1] {\retrievertable};
        \addlegendentry{#1};
    }
    \pgfplotsinvokeforeach{Contriever-MSMARCO}{
        \addplot [bar width=0.07in,bar shift=-0.2in,fill=blue,draw=blue,fill opacity=0.6,draw opacity=0]
            table [y expr=\coordindex, x=#1] {\retrievertable};
        \addlegendentry{#1};
    }
  \end{axis}
  \end{tikzpicture}
  \caption{Retriever performance (Acc@100) without any interventions on target domain corpus}
\label{fig:retriver_zero_shot}
\end{figure}

\pgfplotsset{compat=1.11,
    /pgfplots/xbar legend/.style={
    /pgfplots/legend image code/.code={%
       \draw[##1,/tikz/.cd, rotate=90,yshift=-0.25em]
        (0cm,0cm) rectangle (3pt,1.5em);},
   },
}
\pgfplotstableread[col sep=space]{plots/reader-zero-shot.dat}{\readertable}
\begin{figure}[ht]
    \small
    \begin{subfigure}[t]{0.45\textwidth}
    \centering 
    \begin{tikzpicture}[scale=0.9]
    \begin{axis}[xbar,
      y=0.85cm,
      xmin=0.1,
      xlabel=,
      ytick=data,
      yticklabels from table={\readertable}{dataset},
      nodes near coords,
      every node near coord/.append style={font=\tiny},
      legend columns=3,
      legend style={at={(0.5,1)},anchor=south,font=\normalsize,
      minimum width=1.68cm}]
    
    \pgfplotsinvokeforeach{Source}{
        \addplot [bar width=0.07in,bar shift=0.1in,fill=olive,draw=olive]
            table [y expr=\coordindex, x=#1] {\readertable};
        \addlegendentry{#1};
    }
    \pgfplotsinvokeforeach{UB-RET}{
        \addplot [bar width=0.07in,bar shift=0.0in,fill=orange,draw=orange]
            table [y expr=\coordindex, x=#1] {\readertable};
        \addlegendentry{#1};
    }
    \pgfplotsinvokeforeach{UB-READ}{
        \addplot [bar width=0.07in,bar shift=-0.1in,fill=blue,draw=blue,fill opacity=0.6,draw opacity=0.6]
            table [y expr=\coordindex, x=#1] {\readertable};
        \addlegendentry{#1};
    }
    \addlegendentry{Source}
    \addlegendentry{UB-READ}
    \addlegendentry{UB-RET}
     \end{axis}
    \end{tikzpicture}

    \end{subfigure}
       \caption{Reader performance on target validation set without any interventions. SearchQA, Quasar-S and Quasar-T do not have gold passage annotations so both upperbound are same. The majority voting baseline on COLIEE is 50.95}
    \label{fig:reader_zero_shot}
   
\end{figure}
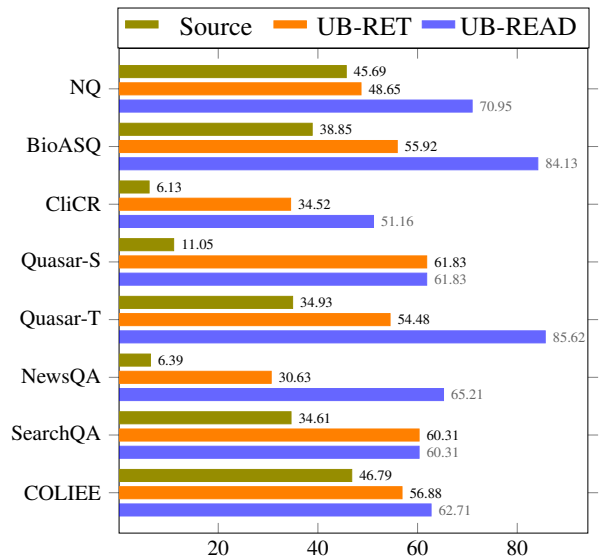


\subsection{Retriever Generalization} To analyze model performance further, we compare the zero-shot generalization performance of four different retrieval models in figure~\ref{fig:retriver_zero_shot}: \textbf{BM25}, \textbf{Contriever}, \textbf{Spider} and \textbf{DPR}.

One observation we find is that Spider, the best performing model on the source domain, exhibits an improvement on SearchQA ($\sim$1\%) (which uses the same underlying source Wikipedia domain), but shows large drops in performance when applied to the target datasets: $\sim$40\% on NewsQA, $\sim$28\% on Quasar-T and, Quasar-S.

To understand the reason for such an enormous performance drop, we sample 50 random incorrect predictions from Spider for manual analysis. We observe two major failure modes. First, we find that dense models are sensitive to changes in the length of contexts. When exposed to documents with heterogeneous lengths that differ from those that they were trained on, models tend to over retrieve shorter contexts. To quantify the sensitivity to changes in lengths on source domains itself, we pool passages from all target corpus into a combined index. We observe that performance of Spider when exposed to this combined index reduces by $\sim$15\% and restricting the minimum length of contexts to be 50 words alleviates the problem and recovers the original performance. The second common failure mode occurs due to changes in distribution of entity types from source to target domain, for instance words like ``plant" in question ``Which is produced in plants of narora kakrapar tarapur" refers to "power plant" in the Wikipedia domain, while in case of Pubmed ``plant" often refers to living organic matter~\cite{sciavolino2021simple}. This is more evident in Spider which uses an auxiliary loss that encourages documents with shared recurring spans (mostly entities) to be closer to each other. This skews model learning to entities seen during training.
Overall, BM25, being an unsupervised method, shows the most competitive performance across all the domains.

    



\section{Interventions for Improving Adaption \label{sec:zero_shot_interventions}}

In the previous section, we hypothesize which target datasets are easily adapted to by a source domain model. Based on the generalizability test, our conjecture was that datasets with a less severe shift like Quasar-S, Quasar-T, and BioASQ would show marked performance improvements with zero-shot adaptation when compared with datasets like CliCR and NewsQA. In the following experiments we observe an average performance improvement of about 8.5\% F1 on datasets with label shift and covariate shift as compared to 3.5\% F1 on datasets with full shift.

\subsection{Zero-shot Adaptation Methods}
\label{sec:zero_shot_adapt}
We perform a series of controlled zero-shot data intervention methods, where we consider the effect of change in distribution of each random variable: question ($\mathbb{Q}$), answer ($\mathbb{A}$) and context ($\mathbb{C}$) one at a time, while keeping the other two fixed.
Our zero-shot interventions utilize only unsupervised (i.e, no question-answer pair annotations) data from the target domain but use source domain data in various ways to generate examples in the target domain. 


\paragraph{Varying context distribution}
To test the effect of change in context distribution, we pool all passages from each dataset into a single document index. 
In figure~\ref{fig:retriver_zero_shot_combined},  we observe that learned models like Spider are sensitive to out-of-domain distractors, especially when a target dataset is based on the source domain corpus (Wikipedia). For instance, SearchQA and NQ both suffer a performance drop of about $\sim$15\%. On the other hand, unsupervised BM25 is much more robust and has a consistent performance even when exposed to a larger pool of documents with the one exception being the legal domain in COLIEE which is a very small index and loses representation when combined with much bugger datasets.

\pgfplotsset{compat=1.9}
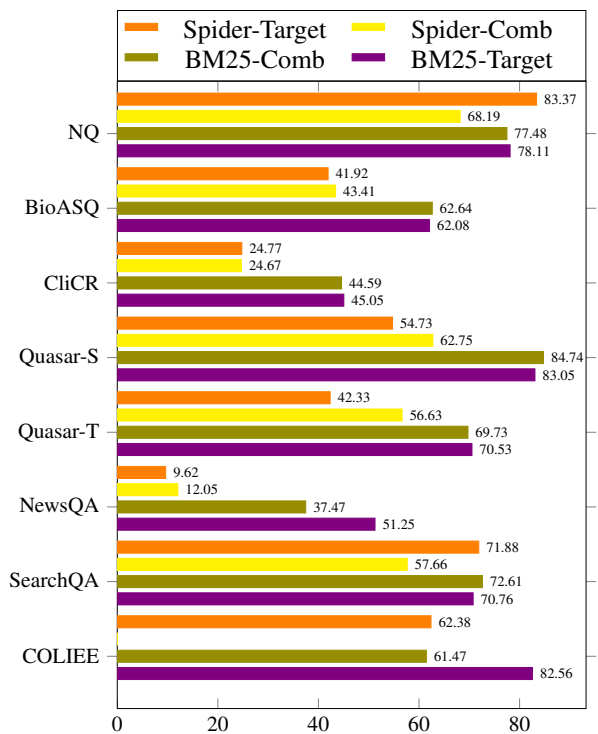
\begin{figure}[ht]
\pgfplotstableread[col sep=space]{plots/retriever-context-distrib.dat}{\retctxtable}
\small
    \centering 
    \begin{tikzpicture}[scale=0.9]
    \begin{axis}[xbar,
      y=1.1cm,
      xmin=0.001,
      ylabel=,
      ytick=data,
      yticklabels from table={\retctxtable}{dataset},
      legend columns=2,
      legend style={at={(0.5,1)},anchor=south,font=\normalsize,
      minimum width=2.83cm},
      nodes near coords,
      nodes near coords style={font=\tiny}    
    ]
      
    \pgfplotsinvokeforeach{Spider-Target}{
        \addplot [bar width=0.07in,bar shift=0.2in,fill=orange,draw=orange]
            table [y expr=\coordindex, x=#1] {\retctxtable};
        \addlegendentry{#1};
    }
    \pgfplotsinvokeforeach{Spider-Comb}{
        \addplot [bar width=0.07in,bar shift=0.1in,fill=yellow,draw=yellow]
            table [y expr=\coordindex, x=#1] {\retctxtable};
        \addlegendentry{#1};
    }
    
    \pgfplotsinvokeforeach{BM25-Comb}{
        \addplot [bar width=0.07in,bar shift=0in,fill=olive,draw=olive]
            table [y expr=\coordindex, x=#1]  {\retctxtable};
        \addlegendentry{#1};
    }
    \pgfplotsinvokeforeach{BM25-Target}{
        \addplot [bar width=0.07in,bar shift=-0.1in,fill=violet,draw=violet]
            table [y expr=\coordindex, x=#1] {\retctxtable};
        \addlegendentry{#1};
    }
    
    \end{axis}
    \end{tikzpicture}
   
    \caption{Retriever Performance (Acc@100): Varying context distribution by creating a combined document index}
    \label{fig:retriver_zero_shot_combined}

\end{figure}

Additionally, in Figure~\ref{fig:reader_ctx_distrib} we show that the FiD reader is not as sensitive as the retriever to changes in context distribution (target vs combined) as we observe only a drop of 3\% in F1 for NewsQA in worst case scenario.

\pgfplotsset{compat=1.3}
\pgfplotstableread[col sep=space]{plots/reader-context-distrib.dat}{\readerctxtable}
\begin{figure}[ht]
    \small
    \begin{subfigure}[t]{0.45\textwidth}
    \centering 
    \begin{tikzpicture}[scale=0.9]
    \begin{axis}[xbar,
      y=0.85cm,
      ylabel=,
      ytick=data,
      xmin=3,
      yticklabels from table={\readerctxtable}{dataset},
      legend columns=3,
      legend style={at={(0.5,1)},anchor=south,font=\normalsize,
      minimum width=1.72cm},
      nodes near coords,
      nodes near coords style={font=\tiny}
      ]
    
    \pgfplotsinvokeforeach{Source}{
        \addplot [bar width=0.07in,bar shift=0.09in,fill=violet,draw=violet]
            table [y expr=\coordindex, x=#1] {\readerctxtable};
        \addlegendentry{#1};
    }
    \pgfplotsinvokeforeach{Target}{
        \addplot [bar width=0.07in,bar shift=0in,fill=teal,draw=teal]
            table [y expr=\coordindex, x=#1] {\readerctxtable};
        \addlegendentry{#1};
    }
    \pgfplotsinvokeforeach{Combined}{
        \addplot [bar width=0.07in,bar shift=-0.09in,fill=olive,draw=olive]
            table [y expr=\coordindex, x=#1] {\readerctxtable};
        \addlegendentry{#1};
    }
    
    \end{axis}
 \end{tikzpicture}

    \end{subfigure}
    \caption{Reader Performance (F1): Effect of change in context distribution with BM25 retrievals from the combined index.}
    \label{fig:reader_ctx_distrib}
\end{figure}
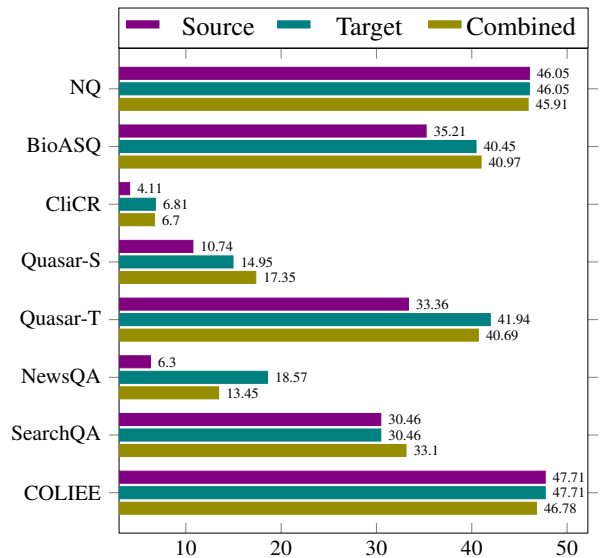

\paragraph{Varying answer distribution}
Many works~\cite{gururangan2018annotation,dua2020benefits,jiang2019avoiding} have shown that unanticipated bias in answer prior distribution can introduce spurious correlations in model learning. In this experiment, we vary the answer distribution by changing the sampling distribution over plausible answer spans. First, we extract and annotate coarse grain entity types from the target corpus using spaCy\footnote{https://spacy.io/}. We then use this coarse-grain entity type information as a set of classes to sample entities to act as cloze-style answers. We choose 50k entities with four different sampling strategies: most frequent, uniformly sampled from entity type categories, randomly sampled from various entity type categories and sampling in proportion to entity type distribution of answers in training set of target dataset.

We choose BioASQ to perform these controlled experiments because the source model has a reasonable end-to-end performance on BioASQ even when retrieving passages from the source domain Wikipedia corpus (Figure~\ref{fig:reader_ctx_distrib}), suggesting that the source corpus contains sufficient information for answering many BioASQ questions. This allows us to use the Wikipedia corpus alone for retrieval, which is useful to control for fixed passage distribution and gauge the impact of the answer distribution in isolation.


In Table~\ref{tab:answer_distrib_ret}, we show that choosing the answer distribution proportional to the uniform distribution across entity type categories boosts retriever performance compared to random sampling, allowing the model to capture all types of answers and generalize better to unseen answer distributions. On the other hand, the best reader model performance is achieved when we know the correct answer distribution of the target dataset upfront, as we see in Table~\ref{tab:answer_distrib_read}. While this demonstrates that answer priors influence reader performance, in an unsupervised setup we will not have this true distribution. Therefore, we adopt the second best technique, i.e., uniform sampling from across the entity type categories for other experiments in the paper (Table~\ref{tab:question_format_read}).

\begin{table}[!htb]
\centering
    \small
    \begin{tabular}{lccc}
    \toprule
      Augmentations &   Acc@100  & \\ 
     \midrule
      Random &  45.35  \\
      Uniform  & 50.02   \\
      Most frequent  & 39.33   \\
      BioASQ train answers & 47.48 \\
    \bottomrule
    \end{tabular}
    \caption{Answer distribution: Retriver performance on BioASQ}
    \label{tab:answer_distrib_ret}
\end{table}

To understand the impact of the pre-trainining vs fine-tuning corpus, we also compare the performance of the FiD reader initialized from T5 pre-trained on common-crawl dataset(C4) compared to one that was pre-trained on pubmed articles. After pretraining, both models are then fine-tuned on our source domain data. In this case, we observe that fine-tuning on a domain that differs from 
 that used in pre-training results in deterioration of model performance.  

\begin{table}[!htb]
\centering
    \small
    \begin{tabular}{lccc}
    \toprule
     Augmentations &  C4 & Pubmed \\ 
     \midrule
      Random  & 33.50 & 33.51 \\
      Uniform & \textbf{39.07} & 35.97 \\
      Most frequent  & \textbf{38.18} & 34.90 \\
      BioASQ train answers & \textbf{41.33} & 36.71 \\
    \bottomrule
    \end{tabular}
    \caption{Answer distribution: Reader performance on BioASQ}
    \label{tab:answer_distrib_read}
\end{table}

\paragraph{Varying question distribution}
To vary the question distribution, we augment the source domain with augmentations generated from the target domain using two different methods. Our first approach uses a question generation~\cite{subramanian2017neural} model trained on the source domain to generate a question given a passage and an answer. This question generation model can be applied to a new target passage and a plausible answer span (entity mention) from the passage~\cite{shakeri2020end,krishna2019generating,song2018leveraging,klein2019learning}. We refer to this method as ``Standard QGen" in table~\ref{tab:quest_format_retriever} and ~\ref{tab:question_format_read}.
Our second approach, which has been less explored previously, converts a sentence in the target corpus to a fill-in-the-blank style cloze question~\cite{taylor1953cloze} by masking a plausible answer span (entity mention) in the sentence. We refer to this method as ``Cloze QA".

In order to sample answers for which we should curate Standard and Cloze QA pairs, we follow the previous subsection and sample answer spans uniformly based on an entity type distribution from the target corpus. We then query our combined index to create a dataset containing cloze style questions aligned with relevant documents. We use these same sampled answers to generate standard QGen QA pairs as well. 


We combine this augmented data with our initial source domain data to train a DPR retriever (Table~\ref{tab:quest_format_retriever}) and a FiD reader (Table~\ref{tab:question_format_read}). We observe similar average performance across both intervention types in retriever and reader models. However, cloze QA pairs are computationally much more efficient to generate as they do not require any additional question generation models.
\begin{table}[!htb]
\centering
    \small
    \begin{tabular}{lccc}
    \toprule
     & { Baseline} & { Cloze QA} & { Standard QGen} \\
     \midrule
      CliCR & 23.87  & \textbf{24.88} & 23.99   \\ 
      BioASQ & \textbf{50.41} & 48.04 & 45.45   \\
      Quasar-S & 50.37  & 66.87 &  \textbf{68.21} \\ 
      Quasar-T & 54.77 & 53.93 &  \textbf{55.57}  \\ 
      NewsQA &  12.54 & \textbf{18.79} &  15.22  \\ 
      SearchQA & \textbf{63.03} &  52.97 & 54.77 \\ 
      COLIEE & \textbf{61.47} & 60.55 & 57.80 \\ 
    \bottomrule
    \end{tabular}
    \caption{Retriever performance: Comparing two types of question formats for augmentation}
    \label{tab:quest_format_retriever}
\end{table}

\begin{table}[tb]
\centering
    \small
    \begin{tabular}{lccc}
    \toprule
      &   Baseline & {Cloze QA} & {Standard QGen}  \\ 
     \midrule
      BioASQ &  45.38 & \textbf{49.41} & 46.43  \\ 
      CliCR &  6.126 & 7.340 & \textbf{10.56} \\
      Quasar-S &  10.24 & \textbf{21.79} &  17.47 \\
      Quasar-T &  34.92 & 41.99 & \textbf{44.73}  \\ 
      NewsQA &  18.57 & \textbf{21.20} & 12.71  \\ 
      SearchQA & 34.60 & \textbf{38.80} & 37.27  \\
      COLIEE & 46.79 & 54.17 &  \textbf{62.38} \\
    \bottomrule
    \end{tabular}
    \caption{Reader performance: Comparing two types of question formats for augmentation}
    \label{tab:question_format_read}
\end{table}
\subsection{Few-shot Generalizability and Adapatability \label{sec:few_shot}}
In section~\ref{sec:zero_shot_interventions}, we saw that zero-shot adaptation does not work well in cases where the target domain distribution is very far from the source domain. As hypothesized in section~\ref{sec:shift}, we would expect improvements from few-shot interventions in the ``Full Shift" datasets NewsQA and CliCR to be more effective than the zero-shot interventions from Section \ref{sec:zero_shot_interventions}. In this section, we find that to be true in addition to the largely across-the-board effectiveness of few-shot interventions.

\paragraph{Few-shot Data Generation}
Zero-shot interventions like question generation models are trained on the source domain and inevitably do not produce generations that are fully compatible with the target domain, leading to degradation when the source and target domains differ drastically. An alternative approach would be to train a question generation model with a few examples from the target domain. However, in practice it is difficult to adapt or fine-tune a question generation and answering model (for validating QA pair correctness) with only a handful of examples. 

\begin{figure*}
\footnotesize
\begin{center}
\small
\begin{tabular}{p{1.3cm}p{10.4cm}p{2.8cm}}
    \toprule
      {\bf Dataset, Corpus} & {\bf Passage} & {\bf Generated Sentence}\\
      \midrule
      BioASQ, Pubmed & Herceptin is widely used in treating Her2-overexpressing breast cancer. However, the application of Herceptin in prostate cancer is still controversial.... This implies that targeting Her2 by both radio- and immunotherapy might be a potential strategy for treating patients with androgen-independent prostate cancer... & Herceptin is a breast cancer drug that has been used in treating prostate cancer. \\
      \hline
      CliCR, Pubmed & An infant was admitted with symptoms of diarrhoea and vomiting. After initial improvement she unexpectedly died. Postmortem confirmed a diagnosis of cytomegalovirus (CMV) enterocolitis. The authors report this case and review other published cases of immunocompetent infants who presented with this infection. Clinicians should consider stool CMV PCR test or referral for endoscopy and biopsy in young babies who present with profuse and prolonged episodes of diarrhoea.
& Immunocompetent infants can present with CMV enterocolitis. \\
      \hline
      Quasar-S, Stackoverflow &  I've recently found scala-bindgen from a Gitter room on Scala Native. Seems like at the present point in time they are developing a tool for generating Scala bindings for C header-files. Are there plans for generating Scala bindings for Objective-C and C++ too... & scala-bindgen -- scala-bindgen is a tool that generates scala bindings for C header files.\\
      \hline
      Quasar-T, Reddit & Interview With Gary James' Interview With Marshall Lytle of Bill Haley's Comets It can be safely said that ``Rock Around The Clock'' was the song by the group Bill Haley And His Comets that started the Rock 'n Roll movement. Still performing today, he spoke about those early days of Rock 'n Roll and his appreciation for what it meant to him.  &  Bill Haley and his comets made rock and roll music \\
      \hline
      NewsQA, CNN/ Dailymail & The Kardashians are already a staple on E! Network . But they've chosen the month of November to assert their dominance on the book world. Kourtney, Kim, and Khloe's first novel,'' Dollhouse ,'' hits shelves today . ``Dollhouse,'' the first fiction endeavor from the Kardashians, follows sisters Kamille, Kassidy, ... & The Kardashians released a new book called 'Dollhouse'.\\
      \hline
      SearchQA, Wikipedia & Charles Henry Dow was an American journalist who co-founded Dow Jones and Company with Edward Jones and Charles Bergstresser. Dow also founded The Wall Street Journal, which has become one of the most respected financial publications in the world... In 1877, he published a History of Steam Navigation between New York and... & Charles Henry Dow, an American journalist, founded The Wall Street Journal in 1882. \\
      \bottomrule
\end{tabular}
\end{center}
\caption{Examples of data generated from few-shot prompting.}
\label{fig:ds}
\end{figure*}

To alleviate this problem, we propose a few shot technique that 
prompts a LLM~\cite{chowdhery2022palm} to generate a sentence given a passage. We use eight seed examples from the target domain to generate additional training data to help bootstrap adaptation in the target domain. We observe that it is easier for large language models to condition on a single variable (context) and compress~\cite{goyal2022news} multiple facts from the passage into a single sentence, as compared to conditioning on a context and answer span together. Moreover, in section~\ref{sec:zero_shot_adapt} we observed that augmentation with cloze style QA pairs yielded similar performance to using question-formatted QA pairs, offering evidence that the precise format is not as important as the content itself. 

We prompt the model in the following format, ``After reading the \emph{article}, <<context>> the \emph{doctor} said <<sentence>>." for pubmed articles.  For other target corpus we replace \emph{doctor} with \emph{engineer}, \emph{journalist} and \emph{poster} for stackoverflow, dailymail and reddit respectively. To filter out invalid sentences, we apply three simple heuristics and remove any generation that 1) includes a number, 2) does not repeat part of the passage verbatim, and 3) has less than 75\% word set overlap with the passage (after removing stopwords). To gauge the precision of our generations, we manually sampled 20 generated sentences for each dataset and found that they were correct more than 70\% of the time.

\begin{table}[!htb]
\centering
    \small
    \begin{tabular}{lcc}
    \toprule
      &   Baseline  & DataGen\\ 
     \midrule
      CliCR &  23.87 &  \textbf{29.06}\\
      BioASQ & 50.41 & \textbf{51.36}\\ 
      Quasar-S & 50.37 & \textbf{71.93} \\ 
      Quasar-T & 54.77 & \textbf{55.47} \\ 
      NewsQA & 12.54 &  \textbf{22.69} \\ 
      SearchQA & 63.03 &\textbf{ 63.35} \\ 
      COLIEE & 73.39 & \textbf{82.23}\\
    \bottomrule
    \end{tabular}
    \caption{Retriever Acc@100 with target specific few shot augmentations (DataGen).}
    \label{tab:ret_few_shot}
\end{table}

To test the retriever performance, we train a DPR model with NaturalQuestions and around $\sim$8k-10k examples, containing pairs of original passage and generated sentence. We compare this model with original source domain DPR model in Table \ref{tab:ret_few_shot}. We observe performance improvements of upto $\sim$18\% in NewsQA and $\sim$21\% in Quasar-S.

\paragraph{Comparison to Few-Shot Closed-Book}
Rather than use a LLM and few-shot prompting to generate data, one could alternatively use the same model and examples to answer questions directly. We next test to what extent the LLM can perform closed-book QA by prompting the same model as used in our data generation with 8 examples that demonstrate how to answer questions in the target domain. In Table \ref{tab:reader_few_shot}, we observe that the LLM does well on datasets with trivia style factual questions, like SearchQA and Quasar-T, but in other cases does not perform as well. The few-shot data augmentation trained model, on the other hand, performs better across a wider range of domains and datasets with the improvements upto $\sim$24\% in F1 on Quasar-S when compared with baseline.

\begin{table}[!htb]
\centering
    \small
    \begin{tabular}{lccc}
    \toprule
      &   Baseline  &  Closed-Book & DataGen \\
     Reader Params & (770M) & (540B) & (770M) \\
     \midrule
      BioASQ & 45.38 & 32.02 & \textbf{50.64}  \\ 
      CliCR &  6.126 & 10.84 & \textbf{19.42}  \\
      Quasar-S &  10.24 & 23.75 & \textbf{34.19}  \\ 
      Quasar-T & 34.92 & \textbf{55.32} & 45.86  \\ 
      NewsQA & 18.57 & 8.67 & \textbf{23.37}  \\ 
      SearchQA & 34.60 & \textbf{61.53} & 37.65  \\
      COLIEE & 46.79 & 53.02 & \textbf{61.11}\\
    \bottomrule
    \end{tabular}
    \caption{Reader: F1 performance with target specific few-shot augmentations (DataGen). Both Close-Book and DataGen use eight examples from the target domain. Few-shot closed-book performance on NQ with eight examples is 36.71}
    \label{tab:reader_few_shot}
\end{table}

In Figure~\ref{fig:ds}, we show qualitative examples generated by our few-shot method depicting that they are able to compose facts from multiple sentences.
\section{Related Work}
Most existing works on domain adaption in question answering consider generalization of the retriever or reader in isolation. 
\paragraph{Domain generalization in readers:} The most popular work in generalization in reading comprehension was introduced as part of the
MRQA~\cite{fisch2019mrqa} challenge, which focuses on transfer of learning from multiple source datasets to unseen target datasets. This multi-task learning setup requires model to perform reasoning at test time that may be unseen at training. It is used as a way to discern what type of reasoning abilities learned at training time are more beneficial for generalization to a cohort of unseen reasoning abilities. However, in this work, we focus on generalization capabilities of an end-to-end ODQA setup to be able read and understand passages in new domain and not the abilities to perform unseen reasoning.

\paragraph{Domain generalization in retrievers:} A recent line of work that test domain generalization of retrievers~\cite{petroni2020kilt,ram2021learning,izacard2022few} focuses on conservative changes to source domain, for instance testing generalization performance of model trained Natural Questions to WebQuestions, TriviaQA -- all of which use the same Wikiepdia corpus. Another line of work follows a recently proposed retrieval bechmark, BEIR \citep{thakur2021beir} that tests generalizability of a general purpose retriever to different corpus/domains. Moreover, it consider only retriever performance in isolation and not end-to-end ODQA performance which can be a brittle metric. Also, it examines the ability of a general purpose retriever to generalize to various domains out-of-the-box and not necessarily how to adapt to a new domain.

\paragraph{Domain adaptation} work in retrievers ~\cite{dai2022promptagator} generate passages in a few shot manner given the query but this does not require the answer (entities) to be correct in the generated passage. \citep{ma2020zero} performs a zero-shot adapatation with noisy labels as it is difficult to train a QA validator in target domain. 
~\cite{siriwardhana2022improving} utilizes examples from target domain in a transfer learning setup while we work in zero to few shot setting.

\section{Conclusion}
In this work we investigated domain generalization in open domain question answering and presented four main contributions. First, we analysed the problems with existing ODQA model and investigate their failure modes. Second, we explored various zero-shot and few-shot data interventions to improve a model's ability to generalize to an unseen target domain. Finally, we described a taxonomy of dataset shift types that provides an way to approximate how effective a source domain trained model can be adapted towards a new target domain.

\bibliography{anthology,custom}
\bibliographystyle{acl_natbib}




\end{document}